\documentclass{ecai2004}
\usepackage{times}
\usepackage{graphicx}
\usepackage{latexsym}

\def\tcsp{\mbox{\em TCSP}}
\def\tcsps{\mbox{\em TCSPs}}
\def\scsp{\mbox{\em SCSP}}
\def\rcc8{\mbox{RCC-8}}
\def\BBR{{\rm I\!R}}
\def\binmat{{\cal B}}
\def\scsps{\mbox{\em SCSPs}}
\def\lbound{\langle ^{\imath}}
\def\rbound{\rangle ^{\jmath}}
\def\modulo{\mbox{ {\em mod} }}
\def\stp{\mbox{\em STP}}
\def\ssp{\mbox{\em SSP}}
\def\bsp{\mbox{\em BSP}}
\def\iaff{\mbox{\em iff}}
\def\lohp{\mbox{ {\em lohp} }}
\def\nhl{\mbox{ {\em nhl} }}
\def\pointr{\mbox{ {\em pt-reg} }}
\def\phl{\mbox{ {\em phl} }}
\def\rohp{\mbox{ {\em rohp} }}
\def\regions{\mbox{ {\em REGIONS} }}
\def\calcul{{\cal CAL}}
\def\lohps{\mbox{{\em lohp}}_{\alpha}}
\def\nhls{\mbox{{\em nhl}}_{\alpha}}
\def\ptregs{\mbox{{\em EQ}}}
\def\phls{\mbox{{\em phl}}_{\alpha}}
\def\rohps{\mbox{{\em rohp}}_{\alpha}}
\def\atoms{\mbox{{\em ATOMS}}}
\def\lohpt{\mbox{{\em lohp}}_{\beta}}
\def\nhlt{\mbox{{\em nhl}}_{\beta}}
\def\phlt{\mbox{{\em phl}}_{\beta}}
\def\rohpt{\mbox{{\em rohp}}_{\beta}}
\def\lhp{\mbox{{\em lhp}}}
\def\rhp{\mbox{{\em rhp}}}
\def\lchps{\mbox{{\em lchp}}_{\alpha}}
\def\rchps{\mbox{{\em rchp}}_{\alpha}}
\def\acwi{\mbox{{\em acwi}}}
\def\bsps{\mbox{\em BSPs}}
\def\holes{\mbox{{\em HOLES}}}
\def\holesplus{\mbox{{\em HOLES}}^+}
\def\queue{{\mbox{\em QUEUE}}}
\begin{document}
\title{\underline{\footnotesize{in Proceedings of the ECAI Workshop on Spatial and
	Temporal Reasoning, pp. 135-139, Valencia, Spain, 2004:}}\\
	A $\tcsp$\thanks{$\tcsps$ stands for Temporal Constraint
	Satisfaction Problems, a well-known constraint-based
	temporal framework \cite{Dechter91a}.}-like decidable
	constraint language generalising existing cardinal
	direction relations}
\author{Amar Isli\\
FB Informatik, Universit\"at Hamburg\\
am99i@yahoo.com}

\maketitle
\bibliographystyle{ecai2004}
\underline{WORK EXACTLY AS REJECTED AT THE MAIN ECAI}\footnote{European
Conference on Artificial Intelligence.}\underline{ 2004}\footnote{The
\underline{reviews} are added to the actual paper, after the references,
for potential people interested in objectivity of conferences' reviewing
processes.}
\begin{abstract}
We define a quantitative constraint language subsuming two calculi
well-known in QSR\footnote{Qualitative Spatial Reasoning.}:
Frank's cone-shaped and projection-based calculi of cardinal direction
relations. The language is based on convex constraints of the form
$(\alpha ,\beta )(x,y)$, with $\alpha ,\beta\in [0,2\pi )$ and
$(\beta -\alpha )\in [0,\pi )$: the meaning of such a constraint is
that point $x$ belongs to the (convex) cone-shaped area rooted at $y$,
and bounded by angles $\alpha$ and $\beta$. The general form of a
constraint is a disjunction of the form
$[(\alpha _1,\beta _1)\vee\cdots\vee (\alpha _{n},\beta _{n})](x,y)$,
with $(\alpha _i,\beta _i)(x,y)$, $i=1\ldots n$, being a convex
constraint as described above: the meaning of such a general constraint
is that, for some $i=1\ldots n$, $(\alpha _i,\beta _i)(x,y)$ holds. A
conjunction of such general constraints is a $\tcsp$-like CSP, which
we will refer to as an $\scsp$ (Spatial Constraint Satisfaction Problem).
We describe how to compute converse, intersection and composition of $\scsp$
constraints, allowing thus to achieve path consistency for an $\scsp$.
We show how to translate a convex constraint into a conjunction of linear
inequalities on variables consisting of the arguments' coordinates.
Our approach to effectively solving a general $\scsp$ is then to adopt a
solution search algorithm using (1) path consistency as the filtering method during the search,
and (2) the Simplex algorithm, guaranteeing completeness, at the
leaves of the search tree.

Keywords: Constraint Satisfaction, Spatial reasoning, Geometric 
Reasoning, Knowledge Representation, Qualitative Reasoning, Quantitative 
Reasoning
\end{abstract}
\newtheorem{df}{Definition}
\newtheorem{thr}{Theorem}
\section{Introduction}\label{sect1}
\begin{flushright}
\begin{tiny}
Conciliating qualitative reasoning and quantitative reasoning in KR\&R systems:\\
a way to systems
	representationally more flexible,
	cognitively more plausible,
	and, 	computationally, with the advantage of having the choice between
		a purely-quantitative and a
		qualitative-computations-first behaviours.
\end{tiny}
\end{flushright}
Knowledge representation (KR) systems allowing for the representation of both
qualitative knowledge and quantitative knowledge are more than needed by modern
applications (see, e.g., \cite{Berleant97a}), which, depending on the level of detail of the knowledge to be
represented, may feel happy with a high-level, qualitative language, or need to use a
low-level, quantitative language. Qualitative languages suffer from what Forbus et al.
\cite{Forbus91a} refer to as the poverty conjecture,
but have the advantage of behaving computationally better.
On the other hand, quantitative languages do not suffer from the poverty conjecture,
but have a slow computatinal behaviour. Thus, such a KR system will feel happier when
the knowledge at hand can be represented in a purely qualitative way, for it can then get rid of heavy numeric calculations, and restrict its computations to a manipulation of symbols, consisting, in the case of constraint-based languages in the style of the Region-Connection Calculus $\rcc8$ \cite{Randell92a}, mainly in computing a closure under a composition table.

An important question raised by the above discussion is clearly how to augment the chances of a qualitative/quantitative KR system to remain at the qualitative level. Consider, for instance, QSR constraint-based, $\rcc8$-like languages. Given the poverty conjecture, which corresponds to the fact that such a language can make only a finite number of distinctions, reflected by the number of its atomic relations, one way of answering the question could be to integrate more than one QSR language within the same KR system. The knowledge at hand is then handled in a quantitative way only in the extreme case when it can be represented by none of the QSR languages which the system integrates.

One way for a KR system, such as described above, to reason about its
knowledge is to start with reasoning about the qualitative part of
the knowledge, which decomposes, say, into n components, one for
each of the QSR languages the system integrates. For $\rcc8$-like
languages, this can be done using a constraint propagation algorithm
such as the one in \cite{Allen83b}. If in either of the n components,
an inconsistency has been detected, then the whole knowledge has been
detected to be inconsistent without the need of going into low-level
details. If no inconsistency has been detected at the high,
qualitative level, then the whole knowledge needs translation into
the unifying quantitative language, and be processed in a purely
quantitative way. But even when the high-level, qualitative
computations fail to detect any inconsistency, they still potentially
help the task of the low-level, purely quantitative computations.
The situation can be compared to standard search algorithms in CSPs,
where a local-consistency preprocessing is applied to the whole
knowledge to potentially reduce the search space, and eventually detect
the knowledge inconsistency, before the actual search for a solution
starts.

With the above
considerations in mind, we consider the integration of Frank's
cone-shaped and projection-based calculi of cardinal direction
relations \cite{Frank92b}, well-known in QSR. A complete decision procedure for the
projection-based calculus is known from Ligozat's work \cite{Ligozat98a}. For the other
calculus, based on a uniform 8-sector partition of the plane, making
it more flexible and cognitively more plausible, no
such procedure is known. For each of the two calculi, the region of
the plane associated with each of the atomic relations is convex,
and given by the intersection of two half-planes. As a
consequence, each such relation can be equivalently written as a
conjunction of linear inequalities on variables consisting of the
coordinates of the relation's arguments. We
consider a more general, qualitative/quantitative language, which, at
the basic level, expresses convex constraints of the form $r(x,y)$, where $r$ is a cone-shaped
or projection-based atomic relation of cardinal directions, or of the
form $(\alpha ,\beta )(x,y)$, with $\alpha ,\beta\in [0,2\pi )$ and
$(\beta -\alpha )\in [0,\pi )$: the meaning of $(\alpha ,\beta )(x,y)$, in particular, is
that point $x$ belongs to the (convex) cone-shaped area rooted at $y$,
and bounded by angles $\alpha$ and $\beta$. We refer to such
constraints as basic constraints: qualitative basic constraint in the
former case, and quantitative basic constraint in the latter. A conjunction of
basic constraints can be solved by first applying constraint
propagation, based on a composition operation to be defined, which is
basically the spatial counterpart of composition of two TCSP
constraints \cite{Dechter91a}. If the propagation detects no
inconsisteny then the knowledge is translated into a system of
linear inequalities, and solved with the well-known Simplex
algorithm. The preprocessing of the qualitative component of the
knowledge can be done with a constraint propagation algorithm such
as the one in \cite{Allen83b}, and needs the composition tables of
the cardinal direction calculi, which can be found in
\cite{Frank92b}.

The general form of a constraint is $(s_1\vee\cdots\vee s_n)(x,y)$, which we
also represent as $\{s_1,\ldots ,s_n\}(x,y)$, where $s_i(x,y)$, for all
$i\in\{1,\ldots ,n\}$, is a basic constraint, either qualitative or
quantitative. The meaning of such a general constraint is that, either
of the $n$ basic constraints is satisfied, i.e.,
$s_1(x,y)\vee\cdots\vee s_n(x,y)$. A general constraint is qualitative if
it is the disjunction of qualitative basic constraints of one type,
cone-shaped or projection-based; it is
quantitative otherwise. The language can be looked at as the spatial
counterpart of Dechter et al.'s TCSPs \cite{Dechter91a}: the domain
of a TCSP variable is $\BBR$, symbolising continuous time, whereas the
domain of an SCSP variable is the cross product $\BBR\times\BBR$,
symbolising the continuous 2-dimensional space.

Due to space limitations, we restrict the presentation to the unifying
$\tcsp$-like constraint language, will all the tools required to convince
the reader that the work is the description of an implementable KR\&R
system, consisting in a search algorithm using path consistency as the
filtering procedure during the search, and the Simplex algorithm as a
completeness guarantee, at the leaves of the search space.
\section{Constraint satisfaction problems}\label{csps}
A constraint satisfaction problem (CSP) of order $n$ consists of:
\begin{enumerate}
  \item a finite set of $n$ variables, $x_1,\ldots ,x_n$;
  \item a set $U$ (called the universe of the problem); and
  \item a set of constraints on values from $U$ which may be assigned to the
    variables.
\end{enumerate}
An $m$-ary constraint is of the form $R(x_{i_1},\cdots ,x_{i_m})$, and asserts
that the values $a_{i_1},\ldots ,a_{i_m}$ assigned to the variables $x_{i_1},\ldots ,x_{i_m}$, respectively,
are so that the $m$-tuple $(a_{i_1},\ldots ,a_{i_m})$ belongs to the $m$-ary relation $R$ (an $m$-ary relation over the
universe $U$ is any subset of $U^m$). An $m$-ary CSP is one of which the
constraints are $m$-ary constraints. We will be concerned exclusively
with binary CSPs.

For any two binary relations $R$ and $S$, $R\cap S$ is the
intersection of $R$ and $S$,
$R\cup S$ is the union of $R$ and $S$, $R\circ S$ is the composition of $R$ and $S$, 
and $R^\smile$ is the converse of $R$; these are defined as follows:
\begin{center}
$
\begin{array}{lll}
R\cap S      &=&\{(a,b):(a,b)\in R\mbox{ and }(a,b)\in S\},\\
R\cup S      &=&\{(a,b):(a,b)\in R\mbox{ or }(a,b)\in S\},\\
R\circ S   &=&\{(a,b):\mbox{for some }c,(a,c)\in R\mbox{ and }(c,b)\in S\},\\
R^\smile     &=&\{(a,b):(b,a)\in R\}.
\end{array}
$
\end{center}
Three special binary relations over a universe $U$ are the empty
relation $\emptyset$ which contains
no pairs at all, the identity relation ${\cal I}_U^b=\{(a,a):a\in U\}$, and the universal
relation $\top _U^b=U\times U$.
\subsection{Constraint matrices}
A binary constraint matrix of order $n$ over $U$ is an $n\times n$-matrix, say $\binmat$,
of binary relations over $U$ verifying the following:
\begin{center}
$
\begin{array}{ll}
(\forall i\leq n)(\binmat _{ii}\subseteq {\cal I}_U^b)   &\mbox{(the diagonal property)},\\
(\forall i,j\leq n)(\binmat _{ij}=(\binmat _{ji})^\smile )&\mbox{(the converse property)}.
\end{array}
$
\end{center}
A binary CSP $P$ of order $n$ over a universe $U$ can be associated with the
following binary constraint matrix, denoted $\binmat ^P$:
\begin{enumerate}
  \item Initialise all entries to the universal relation:
    $(\forall i,j\leq n)((\binmat ^P)_{ij}\leftarrow \top _U^b)$
  \item Initialise the diagonal elements to the identity relation:\\
    $(\forall i\leq n)((\binmat ^P)_{ii}\leftarrow {\cal I}_U^b)$
  \item For all pairs $(x_i,x_j)$ of variables on which a
    constraint $(x_i,x_j)\in R$ is specified:
    $(\binmat ^P)_{ij}\leftarrow (\binmat ^P)_{ij}\cap R,(\binmat ^P)_{ji}\leftarrow ((\binmat ^P)_{ij})^\smile$.
\end{enumerate}
\subsection{Strong $k$-consistency, refinement}
Let $P$ be a CSP of order $n$, $V$ its set of variables and $U$ its universe.
An instantiation of $P$ is any $n$-tuple $(a_1,a_2,\ldots ,a_n)$ of $U^n$,
representing an assignment of a value to each variable.  A consistent instantiation
is an instantiation $(a_1,a_2,\ldots ,a_n)$ which is a solution:
$(\forall i,j\leq n)((a_i,a_j)\in (\binmat ^P)_{ij})$.
$P$ is consistent if it has at least one solution; it is inconsistent otherwise. The
consistency problem of $P$ is the problem of verifying whether $P$ is consistent.

Let $V'=\{x_{i_1},\ldots ,x_{i_j}\}$ be a subset of $V$. The sub-CSP of $P$ generated
by $V'$, denoted $P_{|V'}$, is the CSP with $V'$ as the set of variables, and whose constraint
matrix is obtained by projecting the constraint matrix of $P$ onto $V'$:
$(\forall k,l\leq j)((\binmat ^{P_{|V'}})_{kl}=(\binmat ^P)_{i_ki_l})$.
$P$ is $k$-consistent \cite{Freuder78a,Freuder82a} (see also \cite{Cooper89a}) if for any subset $V'$ of $V$
containing $k-1$ variables, and for any variable $X\in V$, every solution to
$P_{|V'}$ can be extended to a solution to $P_{|V'\cup\{X\}}$. $P$ is strongly
$k$-consistent if it is $j$-consistent, for all $j\leq k$.

$1$-consistency, $2$-consistency and $3$-consistency correspond to node-consistency,
arc-consistency and path-consistency, respectively \cite{Mackworth77a,Montanari74a}.
Strong $n$-consistency of $P$ corresponds to what is called global consistency in
\cite{Dechter92a}. Global consistency facilitates the important task of searching
for a solution, which can be done, when the property is met, without backtracking
\cite{Freuder82a}.

A refinement of $P$ is a CSP $P'$ with the same set of variables, and such that:
$(\forall i,j)((\binmat ^{P'})_{ij}\subseteq (\binmat ^P)_{ij})$.
\section{A spatial counterpart of $\tcsps$: Spatial Constraint Satisfaction Problems ($\scsps$)}
TCSPs (Temporal Constraint Satisfaction Problems) is a constraint-based framewrok
well-known in Temporal Reasoning \cite{Dechter91a}. We provide a spatial
counterpart of $\tcsps$, which we refer to as $\scsps$
---Spatial Constraint Satisfaction Problems. The domain of an $\scsp$ variable is the cross
product $\BBR\times\BBR$, which we look at as the set of points of the 2-dimensional space. As
for a $\tcsp$, an $\scsp$ will have unary constraints and binary constraints, and unary
constraints can be interpreted as special binary constraints by choosing an origin of the
2-dimensional space ---space $(0,0)$.

\begin{figure}
\centerline{\includegraphics[height=5cm]{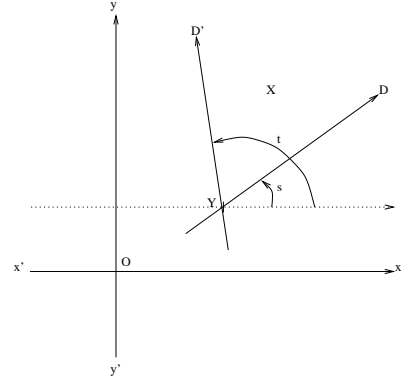}}
\caption{Graphical interpretation of the basic constraint $\lbound s,t\rbound (X,Y)$:
	Given $Y$, the set of points $X$ satisfying the constraint
	$\lbound s,t\rbound (X,Y)$ is the cone-shaped area centred at $Y$, whose
	lower bound (open if $\imath =0$, close otherwise) and upper bound (open if
	$\jmath =0$, close otherwise) are, respectively, the half-lines whose
	angular distances from the $x$-axis, with respect to anticlockwise
	orientation, are $s$ and $t$.}\label{basicconstraint}
\end{figure}

We first define some more terminology to be used in the rest of the paper.
We make use of a Cartesian system of coordinates $(O,x'x,y'y)$. The $x$-axis $x'x$
is the origin of angles, and the anticlockwise orientation is the positive
orientation for angles. Given that we use the set $[0,2\pi )$ as the universe
of angles (measured in radians), if two angles $\alpha$ and $\beta$ are so that $\alpha >\beta$, the
interval $\lbound\alpha ,\beta\rbound$ will represent the union
$\lbound\alpha ,2\pi )\cup [0,\beta\rbound$. Given a positive real number $\alpha$ and
a strictly positive integer $n$, we denote by $\alpha\modulo n$ the remainder
of the integral division of $\alpha$ by $n$. Furthermore, given any
$\alpha ,\beta\in [0,2\pi )$, the difference $\beta\ominus\alpha$ will measure the
anticlockwise (angular) distance of $\beta$ relative to $\alpha$: i.e.,
$\beta\ominus\alpha =(\frac{\beta -\alpha +2\pi}{\pi}\modulo 2)\pi$; similarly, the sum
$\alpha\oplus\beta$ of $\alpha$ and $\beta$ is defined as
$\alpha\oplus\beta =(\frac{\alpha +\beta}{\pi}\modulo 2)\pi$.
\begin{df}[$\scsp$]
An $\scsp$ consists of (1) a finite number of variables ranging over the
universe of points of the 2-dimensional space (henceforth 2D-points); and (2) $\scsp$
constraints on the variables.
\end{df}
An $\scsp$ constraint is either unary or binary, and either basic or disjunctive.
A basic constraint is
(1) of the form $e(x,y)$, $e$ being equality,
or (2) of the general form $\lbound\alpha ,\beta\rbound (x,y)$ (binary) or
$\lbound\alpha ,\beta\rbound (x)$ (unary), with $\alpha ,\beta\in [0,2\pi )$,
$(\beta\ominus\alpha )\in [0,\pi )$, $\imath ,\jmath\in\{0,1\}$.
 `$\langle ^0$' and `$\langle ^1$' stand, respectively, for the left open bracket
`(' and the left close bracket `['. Similarly,
`$\rangle ^0$' and `$\rangle ^1$' stand, respectively, for the right open bracket `)'
and the right close bracket `]'. A graphical illustration of a general basic
constraint is provided in Figure \ref{basicconstraint}.

A disjunctive constraint is of the form
$[S_1
			   \vee\cdots\vee
			       S_n](x,y)$
(binary) or
$[S_1
			   \vee\cdots\vee
			       S_n](x)$
(unary),
with $S_k(x,y)$ and
$S_k(x)$,
$k=1\ldots n$, being basic constraints as described above: in the binary case, the
meaning of such a disjunctive constraint is that, for some $k=1\ldots n$,
$S_k(x,y)$ holds;
similarly, in the unary case, the
meaning is that, for some $k=1\ldots n$,
$S_k(x)$ holds.
A unary constraint $R(x)$ may 
be seen as a special binary constraint if we consider an origin of the 
World (space $(0,0)$), represented, say, by a variable $x_0$: $R(x)$ is
then equivalent to $R(x,x_0)$. Unless explicitly stated otherwise, we
assume, in the rest of the paper, that the constraints of an $\scsp$
are all binary.

An $\scsp$ constraint $R(x,y)$ is convex if, given an instantiation
$y=a$ of $y$, the set of points $x$ satisfying $R(x,a)$ is a convex
subset of the plane. A universal $\scsp$ constraint is an $\scsp$
constraint of the form $[0,2\pi )(x,y)$: the knowledge consisting
of such a constraint is equivalent to ``no knowledge'', i.e., any
instantiation $(a,b)$ of the pair $(x,y)$ satisfies it. A universal
constraint is also a convex constraint. A convex $\scsp$
is an $\scsp$ of which all the constraints are convex.
Given its similarity with an $\stp$ (Simple Temporal Problem)
\cite{Dechter91a}, we refer to a convex $\scsp$ as an $\ssp$ (Simple
Spatial Problem). An $\scsp$ is basic if all its constraints are
basic. We refer to a basic $\scsp$ as a $\bsp$ (Basic Spatial Problem).
Note that a $\bsp$ may have pairs  $(x,y)$ of variables on which no
constraint is specified (the implicit constraint on such pairs is then
the universal relation $[0,\pi )$, which we also refer to as $?$).

The standard path consistency procedure for binary CSPs is guided by three algebraic
operations, the converse of a constraint, the composition of two constraints, and
the intersection of two constraints. These are defined below for $\scsp$ basic
constraints. The case of general (possibly disjunctive) constraints is obtainable
from the case of basic constraints.
\subsection{The converse of an $\scsp$ basic constraint}
The converse of an $\scsp$ relation $R$ is the $\scsp$ relation $R^\smile$ such
that, for all $x$, $y$, $R(x,y)$ $\iaff$ $R^\smile (y,x)$. We refer to the
constraint $R^\smile (y,x)$ as the converse of the constraint $R(x,y)$. The
converse of $e(x,y)$ is clearly $e(y,x)$. The converse of an $\scsp$
basic constraint $\lbound\alpha ,\beta\rbound (x,y)$ is the $\scsp$ basic
constraint $\lbound\alpha\oplus\pi ,\beta\oplus\pi\rbound (y,x)$, which can be explained by
the simple fact that, given any instantiation $(x,y)=(a,b)$ of the pair $(x,y)$
satisfying the constraint $\lbound\alpha ,\beta\rbound (x,y)$, the angle formed by
the directed line $(ba)$ with the $x$-axis is obtained by adding $\pi$ to the angle
formed by the directed line $(ab)$ with the $x$-axis.
\subsection{The composition of two $\scsp$ basic constraints}
Consider a point $y$ of the plane, and an angle $\alpha $ in $[0,\pi )$.
We denote by $l(y,\alpha )$ the directed line through $y$ forming angle
$\alpha $ with the $x$-axis $x'x$. $y$ and $\alpha $ partition the plane into
five zones, which are the left open half-plane bounded by $l(y,\alpha )$,
the half-line consisting of the points of $l(y,\alpha )$ coming before
$y$ (negative half-line), the point $y$ itself, the half-line
consisting of the points of $l(y,\alpha )$ coming after $y$ (positive
half-line), and the right open hal-plane bounded by $l(y,\alpha )$. We
denote the five regions by $\lohp (y,\alpha )$, $\nhl (y,\alpha )$,
$\pointr (y,\alpha )$, $\phl (y,\alpha )$, and $\rohp (y,\alpha )$, respectively,
and the set of all of them by $\regions (y,\alpha )$. Given a fixed angle
$\alpha $ in $[0,\pi )$, we can thus define a five-atom calculus $\calcul _\alpha $ of binary relations.
The atoms are $\lohps$, $\nhls$, $\ptregs$, $\phls$ and
$\rohps$, defined as follows, for all pairs $(x,y)$ of 2D
points: $\lohps (x,y)$ $\iaff$ $x$ belongs to $\lohp (y,\alpha )$,
$\nhls (x,y)$ $\iaff$ $x$ belongs to $\nhl (y,\alpha )$, $\ptregs (x,y)$ $\iaff$
$x=y$, $\phls (x,y)$ $\iaff$ $x$ belongs to $\phl (y,\alpha )$, and
$\rohps (x,y)$ $\iaff$ $x$ belongs to $\rohp (y,\alpha )$. We denote by
$\atoms (\alpha )$ the set of all five atoms. Clearly, $\lohps (x,y)$ $\iaff$
$\rohps (y,x)$, $\nhls (x,y)$ $\iaff$ $\phls (y,x)$, and
$\ptregs (x,y)$ $\iaff$ $\ptregs (y,x)$. In other words, $\lohps$ and
$\rohps$ are each other's converses, and so are $\nhls$ and $\phls$;
whereas $\ptregs$ is its own converse. We consider now two fixed
angles $\alpha $ and $\beta $ from $[0,\pi )$ and compute the composition
$R_1\circ R_2$ of $R_1$ and $R_2$, with $R_1\in\atoms (\alpha )$ and
$R_2\in\atoms (\beta )$. $R_1\circ R_2$ is the relation $R=\{(x,z):
\mbox{ for some $y$, $R_1(x,y)$ and }R_2(y,z)\}$. Clearly, if $R_1$
is $\ptregs$ then $R_1\circ R_2=R_2$, and if $R_2=\ptregs$ then
$R_1\circ R_2=R_1$. We use the standard notation for
(possibly) disjunctive relations. The other possibilities are presented in the
(composition) table of Figure \ref{alphacircbeta}(Top), where:
\begin{figure}
\begin{center}
$
\begin{array}{|l||l|l|l|l|}  \hline
\circ    	&\lohpt    	&\nhlt  &\phlt  &\rohpt\\  \hline\hline
\lohps	 	&ct_0		&ct_1		&ct_2		&?		\\  \hline
\nhls		&ct_3		&ct_4		&ct_5		&ct_6		\\  \hline
\phls		&ct_1		&ct_7		&ct_8		&ct_9		\\  \hline
\rohps		&?		&ct_a		&ct_b		&ct_c		\\  \hline
\end{array}
$
$
\begin{array}{|l||l|}  \hline
\lbound\alpha ,\beta\rbound\mbox{ s.t. }	&\mbox{Translation of }\lbound\alpha ,\beta\rbound (x,y)\  \\  \hline\hline
\alpha\in [0,\pi ),\beta\in [0,\pi )		&(\langle\lhp _{\alpha}\rangle ^{\imath}
				\cap\langle\rhp _{\beta}\rangle ^{\jmath})(x,y)  \\  \hline	 
\alpha\in [0,\pi ),\beta\in [\pi ,2\pi )	&(\langle\lhp _{\alpha}\rangle ^{\imath}
				\cap\langle\lhp_{\beta -\pi}\rangle ^{\jmath})(x,y)  \\  \hline
\alpha\in [\pi,2\pi ),\beta\in [\pi ,2\pi )	&(\langle\rhp _{\alpha -\pi}\rangle ^{\imath}
				\cap\langle\lhp _{\beta -\pi}\rangle ^{\jmath})(x,y)  \\  \hline
\alpha\in [\pi2 ,\pi ),\beta\in [0,\pi )	&(\langle\rhp _{\alpha -\pi}\rangle ^{\imath}
				\cap\langle\rhp _{\beta}\rangle ^{\jmath})(x,y)  \\  \hline
\end{array}
$
\caption{(Top) Composition $R\circ S$, with $R$ atom of $\calcul _\alpha $ and $S$
atom of $\calcul _\beta $. (Bottom) Translation of basic relation $\lbound\alpha ,\beta\rbound$
into $R\cap S$, with $R$ (possibly disjunctive) $\calcul _\alpha $ relation, and
$S$ (possibly disjunctive) $\calcul _\beta$ relation.}\label{alphacircbeta}
\end{center}
\end{figure}
\begin{enumerate}
  \item[$\bullet$] $ct_0$ is $\lohps$ if $\alpha =\beta $, $?$ otherwise;
  \item[$\bullet$] $ct_1$ is $\lohpt$ if $\alpha \geq \beta $, $?$ otherwise;
  \item[$\bullet$] $ct_2$ is $\lohps$ if $\alpha \leq \beta $, $?$ otherwise;
  \item[$\bullet$] $ct_3$ is $\lohpt$ if $\alpha \leq \beta $, $?$ othrwise;
  \item[$\bullet$] $ct_4$ is $\lohps\cap\rohpt$ if $\alpha >\beta $, $\nhls$ if $\alpha =\beta $, $\rohps\cap\lohpt$ otherwise;
  \item[$\bullet$] $ct_5$ is $\lohps\cap\lohpt$ if $\alpha <\beta $, $?_\alpha $ if $\alpha =\beta $, $\rohps\cap\rohpt$ otherwise;
  \item[$\bullet$] $ct_6$ is $\rohpt$ if $\alpha \geq \beta $, $?$ otherwise;
  \item[$\bullet$] $ct_7$ is $\lohps\cap\lohpt$ if $\alpha >\beta $, $?_\alpha $ if $\alpha =\beta $, $\rohps\cap\rohpt$ otherwise;
  \item[$\bullet$] $ct_8$ is $\lohps\cap\rohpt$ is $\alpha <\beta $, $\phls$ if $\alpha =\beta $, $\rohps\cap\lohpt$ otherwise;
  \item[$\bullet$] $ct_9$ is $\rohpt$ if $\alpha <\beta $, $\phls$ is $\alpha =\beta $, $?$ othrwise;
  \item[$\bullet$] $ct_a$ is $\rohps$ if $\alpha \leq \beta $, $?$ otherwise;
  \item[$\bullet$] $ct_b$ is $\rohps$ if $\alpha \geq \beta $, $?$ otherwise;
  \item[$\bullet$] $ct_c$ is $\rohps$ if $\alpha =\beta $, $?$ otherwise;
  \item[$\bullet$] $?	=	\{(p,q):\mbox{ $p$ and $q$ planar points}\}$
		(i.e., $?$ is the universal binary relation on 2D points);
  \item[$\bullet$] $?_\alpha 	=	\{(x,y)\in ?:\mbox{ }x\in l(y,\alpha )\}=\{\nhls ,\ptregs ,\phls\}$
		(i.e., the set of pairs $(x,y)$ of 2D points s.t. $x\in l(y,\alpha )$).
\end{enumerate}
It follows from the above that, given $\alpha ,\beta \in [0,\pi )$, the composition $R_1\circ R_2$
of $R_1\in\atoms (\alpha )$ and $R_2\in\atoms (\beta )$ is a convex relation.
\begin{figure*}
\begin{center}
$
\begin{array}{|l||l|l|l|l|}  \hline
	&\alpha =0	&0<\alpha <\frac{\pi}{2}	&\alpha =\frac{\pi}{2}	&\frac{\pi}{2}<\alpha <\pi\\  \hline\hline
\lohps (X,Y)&y_X>y_Y&y_X-y_Y>tg\alpha .(x_X-x_Y)&y_X<y_Y&y_X-y_Y>tg(\pi -\alpha ).(x_X-x_Y)\\  \hline
\lchps (X,Y)&y_X\geq y_Y&y_X-y_Y\geq tg\alpha .(x_X-x_Y)&y_X\leq y_Y&y_X-y_Y\geq tg(\pi -\alpha ).(x_X-x_Y)\\  \hline
\rohps (X,Y)&y_X<y_Y&y_X-y_Y<tg\alpha .(x_X-x_Y)&y_X>y_Y&y_X-y_Y<tg(\pi -\alpha ).(x_X-x_Y)\\  \hline
\rchps (X,Y)&y_X\leq y_Y&y_X-y_Y\leq tg\alpha .(x_X-x_Y)&y_X\geq y_Y&y_X-y_Y\leq tg(\pi -\alpha ).(x_X-x_Y)\\  \hline
\end{array}
$
\caption{Translation of an $\scsp$ basic constraint into a conjunction of linear inequalities.}\label{scspinequalities}
\end{center}
\end{figure*}

It is now easy to derive the composition, $R\circ S$, of two $\scsp$ basic
constraints $R=\langle ^{\imath _1}\alpha ,\beta\rangle ^{\jmath _1}$ and
$S=\langle ^{\imath _2}\gamma ,\delta\rangle ^{\jmath _2}$. It is sufficient
to know how to translate an $\scsp$ basic relation $R=\lbound\alpha ,\beta\rbound$ into
a conjunction $R_1\cap R_2$, where $R_1$ is a $\calcul _{\alpha}$ or a
$\calcul _{\pi -\alpha}$ convex relation, and $R_2$ a $\calcul _{\beta}$ or a
$\calcul _{\pi -\beta}$ convex relation: this is done in the table of Figure
\ref{alphacircbeta}(Bottom), where the following notation is used.
Given $\alpha\in [0,\pi )$, we denote by $\lchps$ (resp. $\rchps$) the disjunctive
relation $\{\lohps ,\nhls ,e,\phls\}$ (resp. $\{\nhls ,e,\phls ,\rohps\}$).
The constraint $\lchps (x,y)$ (resp. $\rchps (x,y)$) means that $x$ belongs to
the Left (resp. Right) Close Half Plane bounded by $l(y,\alpha )$.
Given $\alpha\in [0,\pi )$ and $\imath\in\{0,1\}$, the notation
$\langle\lhp _{\alpha}\rangle ^{\imath}$ (resp. $\langle\rhp _{\alpha}\rangle ^{\imath}$)
stands for $\lohps$ (resp. $\rohps$) if $\imath =0$, and for $\lchps$ (resp.
$\rchps$) if $\imath =1$. It is important to keep in mind, when reading the table of Figure \ref{alphacircbeta}(Bottom), that $\alpha\in[\pi ,2\pi )$ implies $(\alpha -\pi )\in [0,\pi )$.

The composition $R\circ S$ of basic constraints
$R=\langle ^{\imath _1}\alpha ,\beta\rangle ^{\jmath _1}$ and
$S=\langle ^{\imath _2}\gamma ,\delta\rangle ^{\jmath _2}$ can thus be
written as
$R\circ S=f(\alpha )\circ f(\gamma )\cap f(\alpha )\circ f(\delta )\cap
f(\beta )\circ f(\gamma )\cap f(\beta )\circ f(\delta )$, where $f(x)$,
for all $x\in\{\alpha ,\beta ,\gamma ,\delta\}$, is a $\calcul _x$ atom
if $x\in [0,\pi )$, and a $\calcul _{\pi -x}$ atom if $x\in [\pi ,2\pi )$.
Given that, for all $\alpha ,\beta \in [0,\pi )$, the composition $R_1\circ R_2$
of $R_1\in\atoms (\alpha )$ and $R_2\in\atoms (\beta )$ is a convex relation, we
infer that the composition of two $\scsp$ basic constraint is an $\scsp$ convex
constraint.
\subsection{The intersection of two $\scsp$ basic constraints}
Clearly, $e\cap e=e$; $e\cap\lbound\alpha ,\beta\rbound =e$ if
$\imath =\jmath =1$; and $e\cap\lbound\alpha ,\beta\rbound =\emptyset$ if
$\imath =0$ or $\jmath =0$.

Given a basic relation $R=\lbound\alpha ,\beta\rbound$ and $\gamma\in [0,2\pi )$,
$\gamma$ is anticlockwisely inside $R$ (notation $\acwi (\gamma ,R)$) $\iaff$
(1) $\gamma =\alpha$ and $\imath =1$; (2) $\gamma =\beta$ and $\jmath =1$; or (3)
$\gamma\not =\alpha$ and $\gamma\not =\beta$ and
$\beta\ominus\alpha =(\gamma\ominus\alpha)+(\beta\ominus\gamma )$.

It is now easy to derive the intersection, $R\cap S$, of two $\scsp$ basic
constraints $R=\langle ^{\imath _1}\alpha ,\beta\rangle ^{\jmath _1}$ and
$S=\langle ^{\imath _2}\gamma ,\delta\rangle ^{\jmath _2}$. If neither of
$\acwi (\alpha ,S)$, $\acwi (\beta ,S)$, $\acwi (\gamma ,R)$ and
$\acwi (\delta ,R)$ holds, then $R\cap S=\emptyset$. Otherwise, the
intersection is nonempty: $R\cap S=\lbound\phi ,\theta\rbound$. If
$\acwi (\alpha ,S)$ then $\phi =\alpha$ and $\imath =\imath _1$, otherwise
$\phi =\gamma$ and $\imath =\imath _2$. If $\acwi (\beta ,S)$ then
$\theta =\beta$ and $\jmath =\jmath _1$, otherwise $\theta =\delta$ and
$\jmath =\jmath _2$. Clearly, if
$R\cap S\not =\emptyset$ then it is a basic constraint.

The converse of an $\scsp$ basic constraint is an $\scsp$ basic constraint.
The composition of two $\scsp$ basic constraints is either a basic
constraint or the universal constraint.
Finally, the intersection of two $\scsp$ basic constraints is an $\scsp$
basic constraint.  Now, the only $\scsp$ constraint that may (implicitly)
appear in a $\bsp$ is, as already alluded to, the universal relation $?$.
Furthermore, the converse of $?$ is $?$, $?\cap ?=?$, $?\circ ?=?$, and,
for all basic relations $R$, $R\cap ?=?\cap R=R$ and
$R\circ ?=?\circ R=?$. This leads to the following theorem.
\begin{thr}
The class of $\bsps$ is closed under path consistency: applying path
consistency to a $\bsp$ either detects inconsistency of the latter, or
leads to a (path consistent) $\bsp$.
\end{thr}
It remains, however, to be proven that path consistency terminates when applied
to a $\bsp$. Furthermore, if path consistency is to be used as the filtering
method during the search for a path consistent $\bsp$ refinement of a
general $\scsp$, then it should also be proven that path consistency
terminates when applied to a general $\scsp$ -it may be worth noting here that
path consistency applied to a general TCSP \cite{Dechter91a} may lead to what is
known as the fragmentation problem \cite{Schwalb97a}. We do this through the
explanation of what we refer to as a ``qualitative behaviour'' of path
consistency when applied to a general $\scsp$.
\subsection{Qualitative behaviour of path consistency}
Let $P$ be a general $\scsp$ and $\holes (P)$ the set of all $\gamma\in [0,2\pi )$
such that there exists a constraint $[S_1\vee\cdots\vee S_n](x,y)$ of $P$ with,
for some $i\in\{1,\ldots ,n\}$, $S_i$ of the form $\lbound\alpha ,\beta\rbound$,
and such that $\gamma\in\{\alpha ,\beta\}$. We also denote by $\holesplus (P)$ the
set $\holes (P)
	\cup\{\alpha\in [0,\pi ):\mbox{ }(\alpha +\pi )\in\holes (P)\}
	\cup\{\alpha\in [\pi ,2\pi ):\mbox{ }(\alpha -\pi )\in\holes (P)\}$.
Given a set $A$, we denote by $|A|$ the cardinality of $A$. Clearly $|\holesplus (P)|\leq 2\times|\holes (P)|$.
The qualitative behaviour comes from properties of the operations of converse,
intersection and composition when applied to $\scsp$ basic constraints. The
intersection $R\cap S$ of two $\scsp$ basic constraints
$R=\langle ^{\imath _1}\alpha ,\beta\rangle ^{\jmath _1}$ and
$S=\langle ^{\imath _2}\gamma ,\delta\rangle ^{\jmath _2}$ is of the form
$\lbound\phi ,\theta\rbound$, with both $\phi$ and $\theta$ in
$\{\alpha ,\beta ,\gamma ,\delta\}$. The converse of an $\scsp$ basic constraint
$\lbound\alpha ,\beta\rbound$ is $\lbound\alpha\oplus\pi ,\beta\oplus\pi\rbound$:
but such an operation will not create new ``holes'', since if a basic
constraint of the form $\lbound\alpha ,\beta\rbound(x,y)$ appears in $P$ then we
would have both $\alpha$ and $\beta$ in $\holes (P)$, and both $\alpha\oplus\pi$
and $\beta\oplus\pi$ in $\holesplus (P)$.

Path consistency using, as usual, a queue $\queue$ where to put edges of $P$
whose label has been updated, would thus, for each edge (i.e., pair of variables) $(x,y)$,
successfully update the label at most  $\holesplus (P)$ times. The number of edges is
bounded by $n^2$, $n$ being the number of variables of $P$. Furthermore, when an
edge is taken from $Queue$ for propagation, $O(n)$ operations of converse,
intersection and composition are performed. This leads to the following theorem
stating termination, and providing a worst-case computational complexity, of path
consistency applied to a general $\scsp$.
\begin{thr}
Applying path consistency to a general $\scsp$ $P$ with $n$ variables terminates in
$O(|\holes (P)|\times n^3)$.
\end{thr}
\subsection{Translating an $\scsp$ basic constraint into a conjunction of linear inequalities}
We now provide a translation of an $\scsp$ basic
constraint into (a conjunction of) linear inequalities. We will then be able to
translate any $\bsp$ into a conjunction of linear inequalities, and
solve it with the well-known Simplex algorithm (see, e.g., \cite{Chvatal83a}). This will give a complete solution search algorithm for general $\scsps$, using path consistency at the internal nodes of the search space, as a filtering
procedure, and the Simplex at the level of the leaves, as a completeness-guaranteeing
procedure (the $\scsp$ at the level of a leaf is a path-consistent $\bsp$, but since we
know nothing about completeness of path-consistency for $\bsps$, we need to translate
into linear inequalities and solve with the Simplex).

Given a point $X$ of the plane, we denote by $(x_X,y_X)$ its coordinates. The
translation of $e(X,Y)$ is obvious:
$x_X-x_Y\leq 0\wedge x_Y-x_X\leq 0\wedge y_X-y_Y\leq 0\wedge y_Y-y_X\leq 0$. For the
translation of a general basic constraint
$\lbound\alpha ,\beta\rbound (X,Y)$, the results reported in the table of Figure
\ref{alphacircbeta}(Bottom) imply that all we need is to show how to represent with
a linear inequality each of the following relations on points $X$ and $Y$, where
$\alpha\in [0,\pi )$:
	$\lohps (X,Y)$;
	$\lchps (X,Y)$;
	$\rohps (X,Y)$; and
	$\rchps (X,Y)$.
We split the study into four cases:
$\alpha =0$,
$0<\alpha <\frac{\pi}{2}$,
$\alpha =\frac{\pi}{2}$,
$\frac{\pi}{2}<\alpha <\pi$.
The result is given by Figure \ref{scspinequalities}, where, given an angle $\alpha$,
$tg\alpha$ denotes the tangent of $\alpha$. We remind the reader that in a system of
linear inequalities, there is a way of turning a strict
inequality into a large one \cite{Chvatal83a}.
\section{Summary}\label{summary}
We have provided a $\tcsp$-like decidable constraint language
for reasoning about relative position of points of the 2-dimensional space. The
language, $\scsps$ (Spatial Constraint Satisfaction Problems), subsumes two existing qualitative calculi of relations of
cardinal directions \cite{Frank92b}, and is particularly suited for applications of
large-scale high-level vision, such as, e.g., satellite-like surveillance of a geographic
area. We have provided all the required tools for the implementation of
the presented work; in particular, the algebraic operations of converse,
intersection and composition, which are needed by path consistency. An adaptation
of a solution search algorithm, such as, e.g., the one in \cite{Ladkin92a} (see also
\cite{Dechter91a}), which would use path consistency as the filtering procedure
during the search, can be used to search for a path consistent $\bsp$ refinement of
an input $\scsp$. But, because we know nothing about completeness of path
consistency for $\bsps$, even when a path consistent $\bsp$ refinement exists, this
does not say anything about consistency of the original $\scsp$. To make the search
complete for $\scsps$, we have proposed to augment it with the Simplex algorithm, by
translating, whenever a leaf of the search space is successfully reached, the
corresponding path consistent $\bsp$ into a conjunction of linear inequalities, which
can be solved with the well-known Simplex algorithm \cite{Chvatal83a}.
\bibliography{}
\begin{center}
THE NOTIFICATION LETTER\\
(as received on 3 May 2004)
\end{center}
Dear Amar Isli:

We regret to inform you that your submission 

  C0689
  A TCSP-like decidable constraint language generalising existing
  cardinal direction relations
  Amar Isli
 
cannot be accepted for inclusion in the ECAI 2004's programme. Due to 
the large number of submitted papers, we are aware that also otherwise 
worthwhile papers had to be excluded. You may then consider submitting 
your contribution to one of the ECAI's workshops, which are still open 
for submission.

In this letter you will find enclosed the referees' comments on your 
paper.

We would very much appreciate your participation in the meeting and 
especially in the discussions. 

Please have a look at the ECAI 2004 website for registration details 
and up-to-date information on workshops and tutorials:

    http://www.dsic.upv.es/ecai2004/

The schedule of the conference sessions will be available in May 2004. 

I thanks you again for submitting to ECAI 2004 and look forward to 
meeting you in Valencia.

Best regards
      
Programme Committee Chair
\begin{center}
REVIEW ONE
\end{center}
----- ECAI 2004 REVIEW SHEET FOR AUTHORS -----

PAPER NR: C0689 

TITLE: A TCSP-like decidable constraint language generalising existing 
cardinal direction relations

1) SUMMARY (please provide brief answers)

- What is/are the main contribution(s) of the paper?

A new spatial reasoning calculus is proposed, combining the 
effectiveness of two (one qualitative, one quantitative) calculi.
It is shown that path consistency is terminating on this calculus and 
this motivates a two stage solution technique where the simplex 
algorithm solves leaves of a path consistent instance.

2) TYPE OF THE PAPER

The paper reports on:

  [ ] Preliminary research

  [X] Mature research, but work still in progress

  [ ] Completed research

The emphasis of the paper is on:

  [X] Applications

  [ ] Methodology

3) GENERAL RATINGS

Please rate the 6 following criteria by, each time, using only 
one of the five following words: BAD, WEAK, FAIR, GOOD, EXCELLENT

3a) Relevance to ECAI: EXCELLENT

3b) Originality: EXCELLENT

3c) Significance, Usefulness: FAIR

3d) Technical soundness: GOOD

3e) References: EXCELLENT

3f) Presentation: FAIR

4) QUALITY OF RESEARCH

4a) Is the research technically sound?

    [X] Yes   [ ] Somewhat   [ ] No

4b) Are technical limitations/difficulties adequately discussed?

    [X] Yes   [ ] Somewhat   [ ] No

4c) Is the approach adequately evaluated?

    [ ] Yes   [X] Somewhat   [ ] No

FOR PAPERS FOCUSING ON APPLICATIONS:

4d) Is the application domain adequately described?

    [ ] Yes   [X] Somewhat   [ ] No

4e) Is the choice of a particular methodology discussed?

    [X] Yes   [ ] Somewhat   [ ] No

FOR PAPERS DESCRIBING A METHODOLOGY:

4f) Is the methodology adequately described?

    [ ] Yes   [ ] Somewhat   [ ] No

4g) Is the application range of the methodology adequately described, 
    e.g. through clear examples of its usage?

    [ ] Yes   [ ] Somewhat   [ ] No

Comments:

This paper suggests an algorithm for solving a class of QSRs.  In this 
case it would be useful to see some indication of problems solved used 
this technique that were hard to solve before.

5) PRESENTATION

5a) Are the title and abstract appropriate? 

    [ ] Yes   [X] Somewhat   [ ] No

5b) Is the paper well-organized? [ ] Yes   [X] Somewhat   [ ] No

5c) Is the paper easy to read and understand? 

    [ ] Yes   [X] Somewhat   [ ] No

5d) Are figures/tables/illustrations sufficient? 

    [X] Yes   [ ] Somewhat   [ ] No

5e) The English is  [X] very good   [ ] acceptable   [ ] dreadful

5f) Is the paper free of typographical/grammatical errors? 

    [X] Yes   [ ] Somewhat   [ ] No

5g) Is the references section complete?

    [X] Yes   [ ] Somewhat   [ ] No

Comments:

This paper is hard to read with much detailed technical content.  The 
introduction which is for a KR paper is perhaps too long.  More clarity 
could have been obtained with more effort in the technical sections.

6) TECHNICAL ASPECTS TO BE DISCUSSED (detailed comments)

- Suggested / required modifications:

A running example of an SCSP constraint network is necessary in this 
paper to illustrate the ideas.  Partricularly the decomposition into 
lohp, rohp etc.,  The results of the path-consistency algorithm could then 
be demonstrated.
It is hard to evaluate how effective path-consistency is at pruning in 
such networks.

- Other comments:

This paper needs to be clearer in exposition and technical details.

\begin{center}
REVIEW TWO
\end{center}
----- ECAI 2004 REVIEW SHEET FOR AUTHORS -----

PAPER NR: C0689 

TITLE: A TCSP-like decidable constraint language generalizing existing 
cardinal direction relations

1) SUMMARY (please provide brief answers)

- What is/are the main contribution(s) of the paper?

The paper describes a formalism for reasoning about directions in a 2D 
space with a fixed frame of reference. Basically it uses angular 
sectors analogous to intervals in the 1D case (the relevant calculus in one 
dimension being the formalism of Temporal constraint networks of 
Dechter,  Meiri and Pearl.)

2) TYPE OF THE PAPER

The paper reports on:

  [ ] Preliminary research

  [X] Mature research, but work still in progress

  [ ] Completed research

The emphasis of the paper is on:

  [ ] Applications

  [X] Methodology

3) GENERAL RATINGS

Please rate the 6 following criteria by, each time, using only 
one of the five following words: BAD, WEAK, FAIR, GOOD, EXCELLENT

3a) Relevance to ECAI: GOOD

3b) Originality: FAIR

3c) Significance, Usefulness: WEAK

3d) Technical soundness: WEAK

3e) References: GOOD

3f) Presentation: FAIR

4) QUALITY OF RESEARCH

4a) Is the research technically sound?

    [ ] Yes   [X] Somewhat   [ ] No

4b) Are technical limitations/difficulties adequately discussed?

    [ ] Yes   [X] Somewhat   [ ] No

4c) Is the approach adequately evaluated?

    [ ] Yes   [X] Somewhat   [ ] No

FOR PAPERS FOCUSING ON APPLICATIONS:

4d) Is the application domain adequately described?

    [ ] Yes   [ ] Somewhat   [ ] No

4e) Is the choice of a particular methodology discussed?

    [ ] Yes   [ ] Somewhat   [ ] No

FOR PAPERS DESCRIBING A METHODOLOGY:

4f) Is the methodology adequately described?

    [ ] Yes   [X] Somewhat   [ ] No

4g) Is the application range of the methodology adequately described, 
    e.g. through clear examples of its usage?

    [ ] Yes   [ ] Somewhat   [X] No

Comments:

I think reference should be made to the existing qualitative versions 
of direction calculi represented by Mitra 2000. In particular, as far as 
composing basic relations is concerned, Mitra's paper (about 
"qualitative" relations) contains a simple description which is still  basically 
valid in the quantitative case considered here.

In this respect, I do not think that the paper gives an adequate 
description of the calculus. The description of composition using the five 
relation calculus is convoluted and could be replaced by a much simpler 
one.

About the composition of two basic constraints:
Using the description of a sector-like constraint  (alpha,beta) in 
terms of the intersection of two half-plane constraints is a nice idea (you 
should give more intuition about that). But the resulting tables of 
Fig. 2 are not very easy to use: you could explain the results in a 
simpler way.

About theorem 1: 
What you prove is that, starting from basic constraints o(or the 
universal constraint) and doing composition and intersection results in the 
same kinds of relations. But this does not mean that you can detect 
inconsistency by path-consistency only. This would be a much deeper result!

5) PRESENTATION

5a) Are the title and abstract appropriate? 

    [X] Yes   [ ] Somewhat   [ ] No

5b) Is the paper well-organized? [X] Yes   [ ] Somewhat   [ ] No

5c) Is the paper easy to read and understand? 

    [ ] Yes   [ ] Somewhat   [X] No

5d) Are figures/tables/illustrations sufficient? 

    [X] Yes   [ ] Somewhat   [ ] No

5e) The English is  [ ] very good   [X] acceptable   [ ] dreadful

5f) Is the paper free of typographical/grammatical errors? 

    [ ] Yes   [X] Somewhat   [ ] No

5g) Is the references section complete?

    [ ] Yes   [ ] Somewhat   [ ] No

Comments:

6) TECHNICAL ASPECTS TO BE DISCUSSED (detailed comments)

- Suggested / required modifications:

I would have liked to recommend the paper for acceptance, because I 
think that it 
does contain some good ideas. Unfortunately, in its present state, it 
is still rather weak, because it does not give a sound description of 
the calculus. 

As a consequence, I recommend rejection of the paper. But I think this 
work is worth a good overhaul which would make it into a valuable 
research contribution.

- Other comments:

Some typos:

On page 1
Column 1, line -4:	"computational" instead of "computatinnal".
       line -3: 	A system "will feel happier"?

On page 2
Column 1, line 6:	"constraints"
       Line 30:	"with"?

Column 2, line -14:	"framework instead of "framewrok"

On page 3

In the caption of Fig. 1, "close" should be replaced by "closed".

Column 2, line 11:	"half plane"
\end{document}